\documentclass[letterpaper, 10 pt, conference]{ieeeconf}  

\IEEEoverridecommandlockouts                              

\usepackage{graphics} 
\usepackage{amsmath} 
\usepackage{amssymb}  
\usepackage{booktabs}
\usepackage{multirow}
\usepackage{siunitx}
\usepackage{graphicx}
\usepackage{bm}
\usepackage{cite}
\usepackage{comment}
\usepackage{subfig} 
\usepackage{makecell}

\makeatletter
\let\NAT@parse\undefined
\makeatother
\usepackage[colorlinks,citecolor=green]{hyperref}

\title{\LARGE \bf
Robust and High-Fidelity 3D Gaussian Splatting: Fusing Pose Priors and Geometry Constraints for Texture-Deficient Outdoor Scenes
}
\author{Meijun Guo$^{1,2}$, Yongliang Shi$^{3}$, Caiyun Liu$^{4}$, Yixiao Feng$^{2}$, Ming Ma$^{3}$, \\Tinghai Yan$^{3}$, Weining Lu$^{2*}$, Bin Liang$^{3}$
\thanks{$^{1}$School of Mechatronical Engineering, Beijing Institute of Technology, $^{2}$Beiing National Research Center for Information Science and Technology, $^{3}$Qiyuan Lab, $^{4}$Peking University.}%
\thanks{$*$ Corresponding author.}%
\thanks{Sponsored by the BNRist project (No. BNR2024TD03003) and Qiyuan Innovation Fund (2023-JCJQ-LA-001-075).}
}

\begin{document}

\maketitle
\thispagestyle{empty}
\pagestyle{empty}

\begin{abstract}
3D Gaussian Splatting (3DGS) has emerged as a key rendering pipeline for digital asset creation due to its balance between efficiency and visual quality. To address the issues of unstable pose estimation and scene representation distortion caused by geometric texture inconsistency in large outdoor scenes with weak or repetitive textures, we approach the problem from two aspects: pose estimation and scene representation. For pose estimation, we leverage LiDAR-IMU Odometry to provide prior poses for cameras in large-scale environments. These prior pose constraints are incorporated into COLMAP’s triangulation process, with pose optimization performed via bundle adjustment. Ensuring consistency between pixel data association and prior poses helps maintain both robustness and accuracy. For scene representation, we introduce normal vector constraints and effective rank regularization to enforce consistency in the direction and shape of Gaussian primitives. These constraints are jointly optimized with the existing photometric loss to enhance the map quality. We evaluate our approach using both public and self-collected datasets. In terms of pose optimization, our method requires only one-third of the time while maintaining accuracy and robustness across both datasets. In terms of scene representation, the results show that our method significantly outperforms conventional 3DGS pipelines. Notably, on self-collected datasets characterized by weak or repetitive textures, our approach demonstrates enhanced visualization capabilities and achieves superior overall performance. Codes and data will be publicly available at \href{https://github.com/justinyeah/normal_shape.git}{https://github.com/justinyeah/normal\_shape.git}.

\end{abstract}

\section{Introduction}
The Real2Sim2Real paradigm has emerged as a transformative framework in robotics, enabling systems to bridge reality and simulation through high-fidelity digital twins\cite{Lou2024RoboGSAP}. 
Recent advances in neural scene representations, particularly 3D Gaussian Splatting (3DGS) \cite{kerbl3Dgaussians}, have significantly accelerated Real2Sim pipelines by enabling efficient rendering and intuitive editing of explicit Gaussian primitives \cite{Li2024RoboGSimAR}. These methods rely on camera poses provided by COLMAP \cite {schoenberger2016sfm}. However, the large-scale outdoor environments often feature weak or repetitive textures. Visual mismatches in such environments can lead to geometric texture inconsistencies, which frequently result in failures in batch pose estimation. These failures hinder the subsequent training of Gaussian Splatting. Furthermore, current 3DGS implementations do not fully account for geometric texture consistency, which often results in less realistic rendering results.

Existing neural reconstruction frameworks, including those using Neural Radiance Fields (NeRFs) \cite{mildenhall2020nerf} or 3DGS, rely on Structure-from-Motion (SfM) pipelines, such as COLMAP \cite{pan2024globalstructurefrommotionrevisited}, for pose initialization. While these methods excel at view synthesis, they produce metrically ambiguous models due to projective scale drift, undermining their use in tasks requiring precise collision detection or spatial reasoning \cite{Wang2022ARM, Lou2024RoboGSAP}. Furthermore, the spherical prior of 3DGS introduces surface roughness and inaccuracies at edges, leading to discrepancies in robot-object interaction between Gaussian-based reconstructions and ground-truth LiDAR scans.

To ensure stable and reliable generation of high-fidelity scenes based on 3DGS, we address the challenges caused by geometric texture inconsistencies in two key aspects: \textbf{pose optimization} and \textbf{scene representation}. \textbf{For pose optimization}, we enhance COLMAP's pose optimization by leveraging the stability and reliability of LIO (LiDAR-Inertial Odometry) in large-scale environments to provide a reliable initial pose for each camera. During COLMAP's triangulation, we incorporate initial pose constraints from Fast-LIO2 \cite{FAST-LIO2WeiXU2022} and relative pose constraints between adjacent cameras. This implicitly functions like RANSAC, ensuring geometric texture consistency and preventing pose estimation failures caused by significant mismatches, such as those arising from road features. \textbf{For scene representation}, in addition to employing photometric loss to enforce texture consistency, we also leverage reliable visual normal estimation \cite{DBLP:journals/corr/abs-2110-04994} as supervision. This is achieved by constraining the alignment between the surface normal represented by the Gaussian sphere and the predicted normal \cite{turkulainen2024dnsplatter}. Furthermore, an effective rank \cite{hyung2024effective} regularization term is introduced to constrain the scale and relative scales among the three axes of the Gaussian sphere, thereby preventing the categorization of planar and needle-like Gaussians together.

We summarize our contributions as follows:
\begin{enumerate}
    \item We extend the COLMAP with a new optimization formulation that can employ prior poses for the triangulation process, incorporating initial and relative error constraints by bundle adjustment to enhance the robustness and computational efficiency of pose optimization.
    \item We enhance geometric texture consistency by introducing normal and shape constraints. The normal-constrained optimization applies anisotropic regularization to Gaussian primitives, promoting coherent surface properties. Shape constraints consider all scale parameters and identify relative scales across the three axes, ensuring accurate geometric representation.
    
    \item We experimentally demonstrate that our method can robustly and stably perform pose estimation and 3D reconstruction tasks in large-scale outdoor environments characterized by repetitive textures. We validate our framework against public and custom benchmarks, showing superior geometric consistency compared to state-of-the-art methods.
\end{enumerate}
\section{Related Works}
\subsection{High Fidelity Scene Reconstruction}
\label{related:3dgs_review}

High-fidelity scene reconstruction involves techniques for reconstructing precise 3D models of real-world environments from 2D images, with applications spanning archaeology, architecture,  medicine, and robotics. Common representations include meshes, point clouds, planes, and light fields. At the same time, neural rendering methods like NeRF \cite{mildenhall2020nerf} and its variants \cite{Barron2021MipNeRFAM, Barron2021MipNeRF3U, Mller2022InstantNG, Barron2023ZipNeRFAG, Li2023NeuralangeloHN, Wang2023F2NeRFFN} use multilayer perceptrons (MLPs) to model scene properties such as color and density at spatial coordinates. Despite their accuracy, NeRF’s implicit representation limits rendering efficiency, training speed, and editability—often requiring indirect spatial distortions \cite{Yuan2022NeRFEditingGE} or re-training \cite{Haque2023InstructNeRF2NeRFE3, Gu2023NerfDiffSV} for scene manipulation. Recent breakthroughs in 3DGS \cite{kerbl3Dgaussians} have redefined explicit scene representation, offering a compelling alternative to NeRFs. By leveraging explicit anisotropic Gaussian primitives, 3DGS combines the computational efficiency of raster rendering with fine-grained geometric control—a critical advantage for digital content creation pipelines. 


\subsection{Metric-Consistent 3DGS Reconstruction}
\label{related:metric_consist}

Metric scene information in 3DGS workflows is typically sourced from LiDAR or monocular depth estimators. Recent LiDAR-3DGS fusion methods leverage SLAM frameworks to initialize Gaussians incrementally from odometry trajectories. Examples include Gaussian-LIC \cite{lang2024gaussian}, which combines LiDAR-inertial-camera odometry with adaptive Gaussian splatting; CG-SLAM \cite{hu2024cg} using depth-guided uncertainty-aware fields for efficient RGB-D SLAM; and GS-LIVO/LIV-GaussMap \cite{hong2025gslivorealtimelidarinertial, Hong2024LIVGaussMapLF}, leveraging hierarchical octrees and sliding-window Gaussians for real-time fusion. 
Despite these advancements, SLAM-coupled frameworks struggle with two key issues in digital content creation: (1) pose drift in unbounded scenes from incremental error accumulation; (2) scalability limitations due to resource-intensive simultaneous localization and mapping. 

In contrast, our approach decouples mapping from real-time localization by exploiting pre-registered LiDAR environments. This enables batch optimization: registered LiDAR point clouds provide metric 3DGS initialization, while Structure-from-Motion (SfM) refines poses via multi-view bundle adjustment.

\subsection{Geometrically Accurate 3DGS}
\label{related:geo_regular}
Recent advances in 3DGS have highlighted the critical role of geometric regularization for accurate geometry reconstruction. Traditional 3DGS pipelines struggle with textureless regions and complex geometries due to their spherical Gaussian priors, prompting several key regularization strategies.

\textbf{Depth/Normal Priors} Works similarly to DN-Splatter \cite{turkulainen2024dnsplatter} and VCR-GauS \cite{Chen2024VCRGauSVC}, which integrate depth supervision and photometric consistency constraints, utilizing pre-trained monocular networks to enforce geometric alignment. These methods use adaptive depth losses and view-consistent normal regularization to address the challenges of ill-posed indoor scenes. 


\textbf{Primitive Regulation}
Methods like SuGaR \cite{Gudon2023SuGaRSG} enforce anisotropic orientation constraints aligned with surfaces for high-quality mesh extraction, while PGSR \cite{Chen_2024} applies planar depth regularization and exposure-aware optimization. Similarly, Gaussian surfels \cite{Dai2024HighqualitySR} collapse z-scales to zero, blending surfel-like planar alignment with Gaussian optimization flexibility. More radical reparameterizations include 2D Gaussian splatting \cite{huang20242d}, geometry-enhanced splatting (GES) \cite{Hamdi2024GESGE}, 3D-Convex splatting \cite{Held20243DConvex} and quadratic Gaussian \cite{zhang2024quadraticgaussiansplattingefficient} redefine the primitives entirely. While geometric adaptations enhance surface modeling, redefining primitives risks backward compatibility with standard Gaussian splatting pipelines, demanding a balance between theoretical gains and practical implementation.

\textbf{Implicit Surface Guidance}
Hybrid approaches combining 3DGS with neural signed distance fields (SDF) have emerged to address ambiguous geometries. GSDF \cite{Yu2024GSDF3M} jointly optimizes SDF and Gaussian branches for structure-aware rendering, while GaussianRoom \cite{Xiang2024GaussianRoomI3} uses SDF fields to guide Gaussian densification and applies edge-aware regularization in textureless regions. 3DGSR \cite{Lyu20243DGSRIS} also aligns the geometry from 3DGS with a SDF field. However, obtaining accurate implicit surfaces is non-trivial as they heavily depend on reliable distance field estimations. 


Our key insight lies in achieving metrological accuracy without disrupting the core advantages of 3DGS. Unlike methods requiring prescriptive shape templates or additional neural fields, we construct geometrically meaningful constraints directly from multi-view consistency. This not only preserves compatibility with existing 3DGS toolchains but also enables the decoupled optimization of camera poses and Gaussian attributes—a crucial feature for integrating with SLAM pipelines, as discussed in Section \ref{sec:method}.
\section{Preliminaries}

\begin{figure*}
    \vspace{1em}
    \centering
    \includegraphics[width=0.82\linewidth]{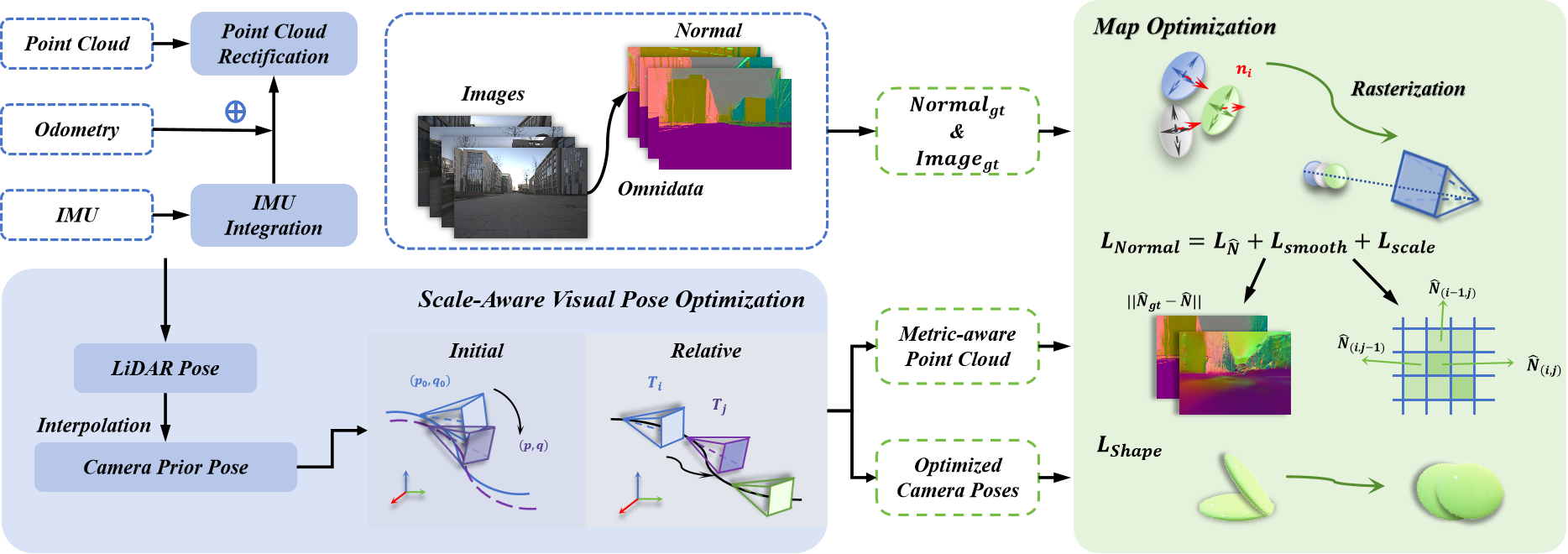}
    \caption{\textbf{Overview of our system.}We utilize LiDAR-Inertial Odometry for initial metric-aware camera poses. Initial and relative constraints refine these poses. By integrating normal and shape constraints, the geometric texture consistency of the scene representation is significantly improved.}
    \label{fig:overview}
\end{figure*}

Our work builds upon 3D Gaussian Splatting (3DGS), a technology that represents a scene using differentiable 3D Gaussian primitives, each parameterized by its mean \( \boldsymbol{\mu} \in \mathbb{R}^{3 \times 3} \), covariance matrix \( \boldsymbol{\Sigma} \in \mathbb{R}^{3 \times 3} \), opacity \( o \in \mathbb{R} \) and view-dependent color \( \boldsymbol{c} \in \mathbb{R}^{3} \) represented by spherical harmonics (SH) coefficients.
The appearance of spatial position $x$ in the scene is contributed by each Gaussian with weight:
\begin{equation}
\label{equa:Gaussian}
\mathcal{G}_{i}\left(\boldsymbol{x} \mid \boldsymbol{\mu}_{i}, \boldsymbol{\Sigma}_{i}\right)=e^{-\frac{1}{2}\left(\boldsymbol{x}-\boldsymbol{\mu}_{i}\right)^{\top} \boldsymbol{\Sigma}_{i}^{-1}\left(\boldsymbol{x}-\boldsymbol{\mu}_{i}\right)}
\end{equation}

Given the world-to-camera transformation $\boldsymbol{W}$, the Jacobian $\boldsymbol{J}$ of the affine approximation to the projective transformation \cite{zwicker2001ewa}, and the camera intrinsic matrix $\boldsymbol{K}$, the 3D Gaussians are projected onto 2D Gaussians for rendering:
\vspace{-0.08in}

\begin{equation}
    \label{equa:projection}
    \begin{aligned}
    \boldsymbol{\mu}_{i}^{\prime} &= \boldsymbol{K} \boldsymbol{W} \left[\boldsymbol{\mu}_{i}, 1\right]^{\top}, 
    \boldsymbol{\Sigma}_{i}^{\prime} = \boldsymbol{J} \boldsymbol{W} \boldsymbol{\Sigma}_{i} \boldsymbol{W}^{\top} \boldsymbol{J}^{\top} \\
    \end{aligned}
\end{equation}

The parameters of the Gaussians are optimized by minimizing a composite loss function that compares the rendered image against the ground truth.
\vspace{-0.03in}

\begin{equation}
    \mathcal{L}_{img}=(1-\lambda_\mathrm{ssim})\mathcal{L}_1+\lambda_\mathrm{ssim}\mathcal{L}_\mathrm{SSIM}
\end{equation}
With adaptive density control (splitting/cloning oversized Gaussians, culling low-opacity primitives) during stochastic gradient descent. 

\section{Method}
\label{sec:method}

As illustrated in Fig.\ref{fig:overview}, our framework integrates a multi-stage reconstruction pipeline. First, LiDAR-Inertial Odometry is used to obtain accurate metric poses. Leveraging lidar-camera extrinsics, we derive camera poses through linear interpolation and spherical linear interpolation (slerp) between consecutive lidar poses. These initial poses enable COLMAP to perform reconstruction while maintaining metric consistency. Subsequently, pose estimation is optimized with both initial and relative pose constraints, resulting in metric-aware point cloud data and refined camera poses.

For reconstruction, we use the refined camera poses and input images. To regularize the geometry, we incorporate normal maps from the Omnidata \cite{DBLP:journals/corr/abs-2110-04994} model as an auxiliary form of supervision. Additionally, we introduce shape constraints to ensure geometric consistency. By optimizing these constraints, we achieve significant improvements in the experimental results, thereby enhancing both the accuracy and reliability of the reconstruction model.

\subsection{Scale-Aware Visual Pose Optimization}
\label{scale_aware_pose}


To achieve metric-scale reconstruction, we use FAST-LIO2 \cite{FAST-LIO2WeiXU2022}, a LiDAR-Inertial Odometry (LIO) framework. LiDAR can provide metric-scale data through direct depth measurements. We synchronize timestamps to compute camera poses via spatio-temporal interpolation, combining linear interpolation for positions and spherical linear interpolation for orientations. This generates metric-aware camera pose priors aligned with the world frame.



While COLMAP is a widely used Structure-from-Motion (SfM) framework, it typically produces sparse 3D reconstructions without metric scale, due to its reliance on relative image constraints. Although foundational for methods like 3DGS and NeRFs, this scale ambiguity compromises physical accuracy in large-scale outdoor scenes, often resulting in integration errors and inaccurate positioning. To address this, we initialize COLMAP optimization using metric-scale camera poses based on FAST-LIO2, thereby enforcing geometric consistency. Feature matching and triangulation are enhanced with metric pose priors and relative constraints, stabilizing convergence and accelerating computation compared to traditional SfM pipelines.

To quantify the deviation between the current pose, consisting of position error $\boldsymbol{e}_p$ and rotation error $\boldsymbol{e}_q$, we compute the difference between the current pose $(\boldsymbol{R}, \boldsymbol{p})$ and the initial pose $(\boldsymbol{R}_0, \boldsymbol{p}_0)$ using Equation.(\ref{equa:pose_t}).
\vspace{-0.05in}
\begin{equation}
    \begin{aligned}
    \label{equa:pose_t}
    \boldsymbol{e}_p & = \boldsymbol{p}_0 - \boldsymbol{p} \\
    \boldsymbol{e}_q & = \mathrm{Log}(\boldsymbol{R}_{0}^{T} \cdot \boldsymbol{R})
    \end{aligned}
\end{equation}
We can perform optimization based on a prior pose constraint, as shown in Equation.(\ref{equa:prior_delta}).
\begin{equation}
    \begin{aligned}
    \label{equa:prior_delta}
     \frac{\partial\boldsymbol{e}_p}{\partial \boldsymbol{p}} &= \frac{\boldsymbol{p}_0 -(\boldsymbol{p} + \delta \boldsymbol{p} )  - \boldsymbol{p}_0 + \boldsymbol{p}}{\delta \boldsymbol{p}} = \boldsymbol{I}
    \\
    \frac{\partial\boldsymbol{e}_q}{\partial \boldsymbol{R}} &= \boldsymbol{J_r^{-1}}(\mathrm{Log}(\boldsymbol{R}_0^T \boldsymbol{R}))
    \end{aligned}
\end{equation}
where $\boldsymbol{J_r^{-1}}(\cdot)$ denotes the inverse of the right Jacobian of $\mathrm{SO}(3)$, which captures the first-order approximation of the Lie algebra perturbation on the manifold.

The relative pose constraints are defined in Equation.(\ref{equa:partial_ti}) and Equation.(\ref{equa:delta_partial}).
\begin{equation}
\label{equa:partial_ti}
    \begin{aligned}
    & \boldsymbol{e}_{intra}  =\mathrm{Log}(\boldsymbol{P}_i^{-1}\boldsymbol{P_j}) \\
    & \quad =
    \begin{bmatrix}
    \mathrm{Log}(\boldsymbol{R}_{ij}^T\boldsymbol{R}_i^T\boldsymbol{R}_j) & \boldsymbol{R}_{ij}^T\boldsymbol{R}_i^T(\boldsymbol{p}_j-\boldsymbol{p}_i)-\boldsymbol{R}_{ij}^T\boldsymbol{p}_{ij} \\
     \\
    \boldsymbol{0} & \boldsymbol{1}
    
    \end{bmatrix}
    \end{aligned}
\end{equation}
where the relative pose error between camera $i$ and $j$ is characterized. The transformation matrix is converted to Lie algebra space, enabling a precise quantification of the disparity between the current and desired poses of the two cameras.
\begin{equation}
\label{equa:delta_partial}
    \begin{aligned}
    & \frac{\partial\boldsymbol{e}_{ij}}{\partial\boldsymbol{p}_i}=
        \begin{bmatrix}
            \boldsymbol{0}_{3\times3} & -\boldsymbol{R}_{ij}^T\boldsymbol{R}_i^T
        \end{bmatrix} \\
    & \frac{\partial\boldsymbol{e}_{ij}}{\partial\boldsymbol{p}_j}=
        \begin{bmatrix}
            \boldsymbol{0}_{3\times3} & \boldsymbol{R}_{ij}^T\boldsymbol{R}_i^T
        \end{bmatrix} \\
    & \frac{\partial\boldsymbol{e}_{ij}}{\partial\boldsymbol{R}_i}=
        \begin{bmatrix}
            -\boldsymbol{J}_r^{-1}(\mathrm{Log}(\boldsymbol{R}_{ij}^T\boldsymbol{R}_i^T\boldsymbol{R}_j))\boldsymbol{R}_j^T\boldsymbol{R_i} & \boldsymbol{R}_{ij}^T(\boldsymbol{R}_i^T(\boldsymbol{p}_j\boldsymbol{-}\boldsymbol{p}_i))^\wedge
        \end{bmatrix} \\
    & \frac{\partial\boldsymbol{e}_{ij}}{\partial\boldsymbol{R}_j}=
        \begin{bmatrix}
        \boldsymbol{J}_r^{-1}(\mathrm{Log}(\boldsymbol{R}_{ij}^T\boldsymbol{R}_i^T\boldsymbol{R}_j)) & \boldsymbol{0}_{3\times3}
        \end{bmatrix}
    \end{aligned}
\end{equation}
where the first two equations correspond to translations $ \boldsymbol{p}_i$ and $\boldsymbol{p}_j$ , while the last two relate to rotations $\boldsymbol{R}_i $ and $\boldsymbol{R}_j$.

\subsection{Map Optimization}
Our optimization seeks to transform the Gaussians into flat disk-like shapes, where the minor axis (approximating the surface normal direction) is significantly smaller than the others. To enforce this planar geometry, we regularize the smallest scale component $\boldsymbol{S}_k = diag(s_{k_1}, s_{k_2}, s_{k_3})$ array all Gaussian minimization $L_1$:
\begin{equation}
    \label{equa:min_scale}
    \mathcal{L}_{\mathrm{scale}}=\sum_{i}\left\|\min \left(s_{k_1}, s_{k_2}, s_{k_3}\right)\right\|_{1}
\end{equation}

To suppress degenerate ``needle-like'' Gaussians during optimization, we enforce ordered scaling axis constraints $ s_{k_1}^2 \ge s_{k_2}^2\ge s_{k_3}^2 > 0 $ and regularize their relative magnitudes via entropy. This ensures the Gaussian resembles a planar surface when observed, preserving numerical stability. We define entropy with normalized axis weights:
\begin{equation}
    \label{equa:En_GS}
    \begin{aligned}
    w_{k_i} &= \frac{{s_{k_i}}^2}{ \sum_{i} {s_{k_i}}^2} \\
    En_k &= \exp \left ( - \sum_{i}{ (w_{k_i} \log(w_{k_i})) }  \right ) 
    \end{aligned}
\end{equation}

Therefore, we define the loss that constrains the Gaussian shape in the Equation. (\ref{equa:L_erank}). When \( En_k = 2 \), the Gaussian shape is optimal, and we set \( \lambda_{En} = 0.01 \).
\vspace{-0.05in}
\begin{equation}
    \label{equa:L_erank}
    \mathcal{L}_{En} = \lambda_{En} \sum_{k}{\max \left (  - \log({En_{k} - 1 + \epsilon }), 0 \right ) }
\end{equation}


\begin{figure}
    \vspace{2em}
    \centering
    \includegraphics[width=0.65\linewidth]{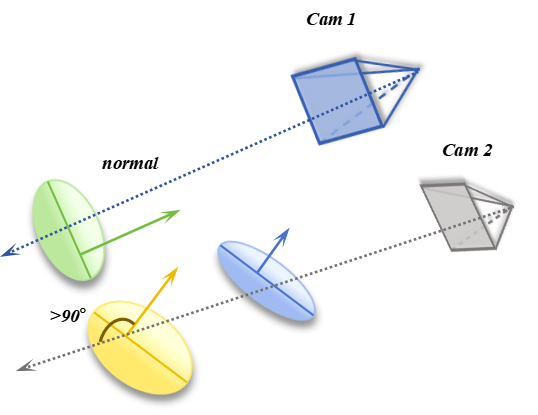}
    \caption{Illustration of the angle between the Gaussian normal and the camera optical axis.}
    \label{fig:normaldegree}
\end{figure}

To further align geometric priors with reconstruction fidelity, we introduce a normal consistency loss. Surface normals directly encode local geometry, and enforcing their accuracy enhances the recovery of fine-grained details. We constrain each Gaussian's minor axis to align with its estimated surface normal. This prevents degenerate ``linear'' projections under certain camera angles Fig. \ref{fig:normaldegree} by imposing an angular threshold: the deviation between a Gaussian's minor axis and the true normal must exceed 90\textdegree{} to penalize misoriented primitives.

Normal maps are rendered via the $\alpha$-blending Equation.(\ref{equa:normal_map}), similar to RGB rendering. 
\vspace{-0.05in}
\begin{equation}
    \label{equa:normal_map}
    \hat{\boldsymbol {N} }= \sum_{i \in \boldsymbol{N}}{ \boldsymbol{n}_i \alpha_i \boldsymbol{T}_i }
\end{equation}
where $\boldsymbol{T}_i$ is the accumulated transmittance at pixel.

These predicted normals are supervised using high-fidelity normal maps $\boldsymbol{N}_{gt}$ from the Omnidata model, which provides robust and smooth geometric priors. The loss minimizes deviations between rendered and ground truth normals using image gradients:
\vspace{-0.05in}

\begin{equation}
    \label{equa:L_normal}
    \mathcal{L}_{\hat{ \boldsymbol{N}}}=\frac{1}{W} \sum_{p \in  \boldsymbol{W}}|\overline{\nabla \boldsymbol{I}}|^{5}\left\| \hat{\boldsymbol{N}}_{gt}-\hat{\boldsymbol{N}}\right\|_{1}
\end{equation}
where $\overline{\nabla \boldsymbol{I}}$ is the ground truth image gradient, which is normalized to the range of 0 to 1, and $\boldsymbol{W}$ is the set of image pixels.


Surface normals should vary smoothly across adjacent regions unless interrupted by genuine geometric discontinuities (e.g., edges or occlusions). To encourage local coherence in geometry, we impose a smoothness loss that regularizes the spatial consistency of predicted normals. We quantify this local continuity by penalizing normal discrepancies in two adjacent directions.
\begin{equation}
    \label{equa:L_smooth}
    \mathcal{L}_{\text {smooth}} = \frac{1}{|\hat{\boldsymbol{N}}|} \sum_{i, j}\left(\left| \hat{\boldsymbol{N}}_{i+1, j}- \hat{\boldsymbol{N}}_{i, j}\right|+\left| \hat{\boldsymbol{N}}_{i, j+1} - \hat{\boldsymbol{N}}_{i, j}\right|\right)
\end{equation}
where $\hat{\boldsymbol{N}}_{i,j}$ is the predicted normal map at the pixel $(i,j)$.

The final loss we use for optimization is defined in Equation. (\ref{equa:optimization}), and we set $\lambda_{smooth}$ = 0.5:
\begin{equation}
\label{equa:optimization}
\mathcal{L}=\mathcal{L}_{\mathrm{img}} + \underbrace{\mathcal{L}_{En}}_{\mathcal{L}_{shape}} + (\underbrace{ \mathcal{L}_{\mathrm{scale}} +\mathcal{L}_{\hat{\boldsymbol{N}}}+\lambda_{\mathrm{smooth}}\mathcal{L}_{\hat{\boldsymbol{N}}_{\mathrm{smooth}}}}_{\mathcal{L}_{\mathrm{normal}}})
\end{equation}
\section{Experiments}


\textbf{Dataset.} Our evaluation targets large-scale real-world outdoor environments, utilizing two datasets to ensure diverse geographic and structural coverage. a) The HKU-Campus (HKU) dataset, sourced from \cite{DBLP:journals/corr/abs-2109-07982}, comprises synchronized LiDAR point clouds and RGB imagery spanning mixed indoor and outdoor scenes. b) We introduce QY, a proprietary dataset captured via a four-wheeled robotic platform equipped with a Mid-360 LiDAR for precise spatial mapping and a FLIR Blackfly camera for high-resolution imagery. The QY dataset features extended trajectories and complex geometric layouts, capturing detailed outdoor structures in a campus environment to address gaps in existing benchmarks.

\textbf{Evaluation Metrics.}
We assess rendering quality through three standard metrics: PSNR and SSIM \cite{1284395}, which evaluate photometric accuracy (higher values preferred), and LPIPS \cite{DBLP:journals/corr/abs-1801-03924}, a re-implementation that adheres to the original 3DGS work.

\textbf{Implementation Details.}
The experiments are conducted on a single NVIDIA GeForce RTX 4060 Ti GPU with 16 GB VRAM. The proposed approach is implemented using PyTorch \cite{DBLP:journals/corr/abs-1912-01703} and the gsplat library (v1.1.1) \cite{ye2024gsplatopensourcelibrarygaussian}, a re-implementation that strictly adheres to the original 3DGS work. Each model undergoes 30,000 training iterations, with Gaussian densification halted at 15,000 iterations to prevent overfitting.


\begin{table}[ht]
\centering
\caption{Rendering Results with Different Constraints}
\label{table:ex_1}
\begin{tabular}{lllS[table-format=1.3]S[table-format=2.3]S[table-format=1.4]}
\toprule
Dataset & ${\mathcal{L}_{\mathrm{normal}}}$ & ${\mathcal{L}_{shape}}$ & {\textbf{LPIPS}$\downarrow$} & {\textbf{PSNR}$\uparrow$} & {\textbf{SSIM}$\uparrow$} \\
\midrule
\multirow{4}{*}{HKU} 
  & $\times$ & $\times$ & 0.197 & 25.806 & 0.8374 \\
  & $\times$ & $\checkmark$ & 0.177 & 27.022 & 0.8502 \\
  & $\checkmark$ & $\times$ & 0.173 & 27.048 & 0.8525 \\
  & $\checkmark$ & $\checkmark$ & \textbf{0.173} & \textbf{27.238} & \textbf{0.8543} \\
\midrule
\multirow{4}{*}{QY} 
  & $\times$ & $\times$ & 0.373 & 25.311 & 0.8436 \\
  & $\times$ & $\checkmark$ & 0.354 & 26.562 & 0.8532 \\
  & $\checkmark$ & $\times$ & 0.353 & 26.638 & 0.8542 \\
  & $\checkmark$ & $\checkmark$ & \textbf{0.353} & \textbf{26.729} & \textbf{0.8543} \\
\bottomrule
\end{tabular}
\end{table}






\textbf{Ablation Study for Scene Representation:} As shown in Table \ref{table:ex_1}, we conducted a series of experiments using ${\mathcal{L}_{\mathrm{normal}}}$ and ${\mathcal{L}_{En}}$ constraints on the HKU and QY, and evaluated the performance using LPIPS, PSNR, and SSIM. Based on our refined poses, adopting both the normal constraint and shape constraint individually outperforms the original 3DGS, while the combination of both geometric constraints achieves the best results.  

\textbf{Pose Optimization Evaluation:} The quantitative results are presented in Table \ref{table:poseevo}. Without pose refinement, using only the camera poses interpolated from LIO along with our complete scene representation, the performance degrades significantly. This highlights the effectiveness and accuracy of our pose optimization method. We also compare our method with R3LIVE\cite{DBLP:journals/corr/abs-2109-07982}, a representative LiDAR-IMU-Camera odometry system. R3LIVE performs worse than our approach, especially on the QY dataset, which features weak and repetitive textures. This failure occurs regardless of whether our scene representation is used. Both quantitatively and visually, our approach outperforms R3LIVE in such challenging scenarios.
Furthermore, COLMAP-PCD is highly sensitive to environmental conditions, often resulting in pose estimation failures on both datasets, particularly in scenes with low texture or geometric ambiguity. On the HKU dataset with rich textures, COLMAP achieves more accurate pose estimation due to its reliance on reliable feature matching, where homogeneous sensors ensure better geometric texture consistency. Our method achieves pose estimation at triple the speed of COLMAP (one-third runtime) while preserving metric-scale accuracy. More critically, in texture-degraded scenarios, such as the QY dataset—where COLMAP fails entirely—our approach consistently achieves robust and accurate pose estimation, demonstrating both computational efficiency and strong resilience to environmental degradation.

\begin{figure*}
    \centering
    \includegraphics[width=0.8\linewidth]{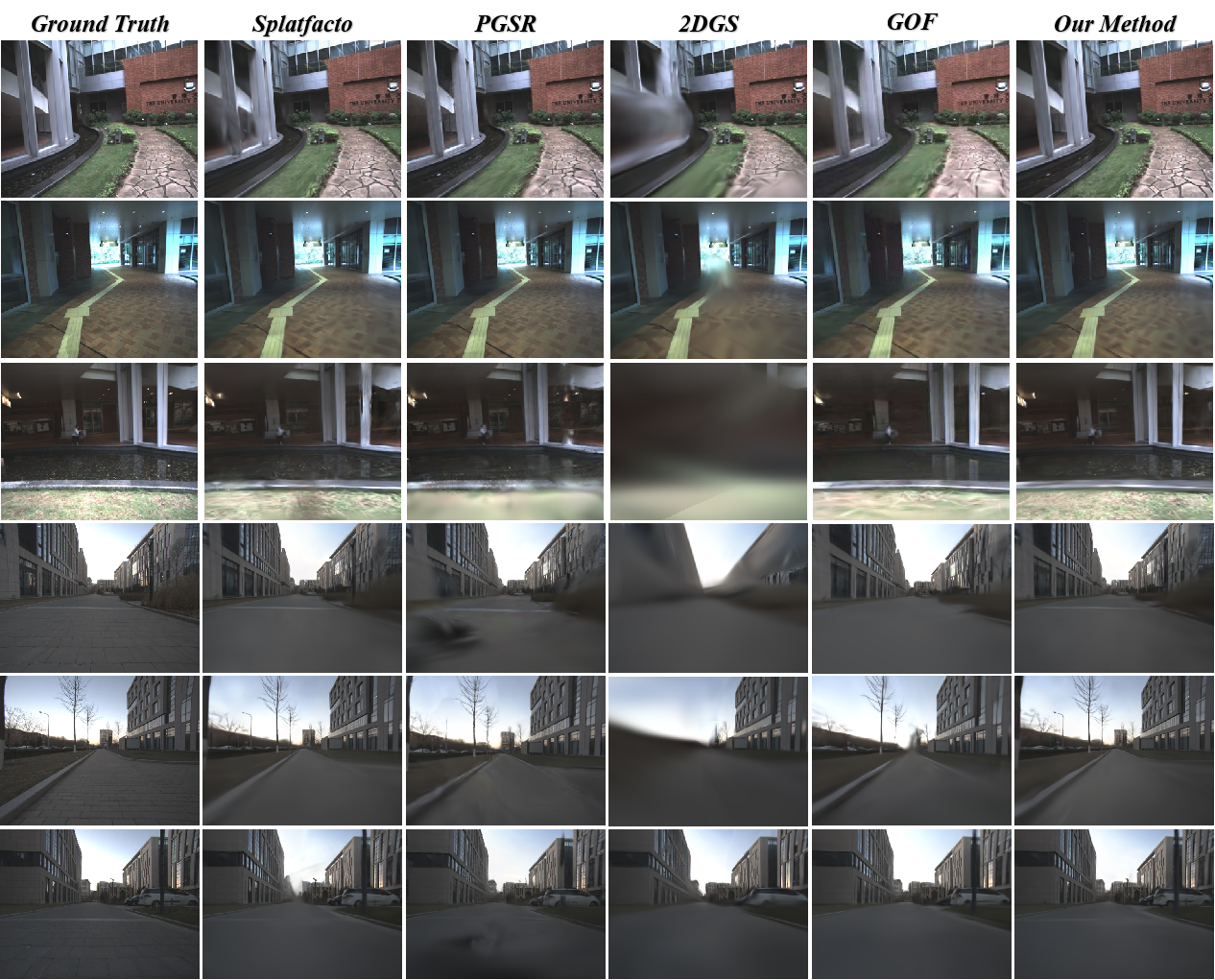}
    \caption{Comparison of Rendering Results.}
    \label{fig:experiment_all}
\end{figure*}

\textbf{Comparison with other baselines: } 
Given the poses obtained from our optimization method, four scene representation baselines are evaluated for comparison. Among these, PGSR \cite{Chen_2024} and 2DGS \cite{huang20242d} are geometric constraint approaches based on surface constraints and regularization techniques, respectively. SplatFacto incorporates the original 3DGS \cite{kerbl3Dgaussians}, while GOF \cite{yu2024gaussianopacityfieldsefficient} represents a typical GS method designed explicitly for unbounded outdoor scenes.
As the Table \ref{table:baseline} depicts, 
While PGSR demonstrates superior overall performance on the QY dataset in quantitative evaluations, the visualization results in Fig.~\ref{fig:experiment_all} indicate that our method reconstructs scene details more effectively, particularly in texture-deficient regions such as lawns and roads. PGSR exhibits sharper edge information due to its exclusive reliance on normal constraints. In contrast, our approach incorporates smoothness and effective rank regularization terms, which enhance rendering continuity and mitigate abrupt pixel-level variations in surrounding areas. Consequently, in weakly textured scenes dominated by edge information, regions lacking texture exhibit minimal photometric variation, making fine-grained quantitative comparisons challenging despite high reconstruction quality. Moreover, quantitative metrics tend to favor methods that preserve sharper edge features. Thus, while PGSR achieves stronger numerical performance on the QY dataset, its rendering results are less photorealistic than those of our method.
Additionally, 2DGS performs well in texture-rich regions but struggles in texture-poor areas. Meanwhile, GOF lacks explicit constraints such as normal vectors and scale consistency, and is susceptible to local ambiguities in repetitive structures. These ambiguities, caused by repeated textures within the opacity field, lead to surface distortions. As a result, both 2DGS and GOF perform suboptimally in both quantitative and visual evaluations.

\begin{table*}[ht]
\vspace{1em}
    \centering
    \caption{Pose estimation evaluation.}
    \label{table:poseevo}
    \begin{tabular}{c c r r r c r r}
        \toprule
        \textbf{Dataset} & \textbf{Metric} &\textbf{\makecell{R3live\\+3DGS}} & \textbf{\makecell{R3live\\+OurGS}} & \textbf{COLMAP} & \textbf{\makecell{COLMAP\\-PCD}} & \textbf{\makecell{w/o pose \\ refinement}} &\textbf{All Ours} \\
        \midrule
        \multirow{3}{*}{HKU}
            & \textbf{LPIPS}$\downarrow$  & 0.191 & 0.195 & \textbf{0.159}& - & 0.283 &0.173 \\
            & \textbf{PSNR}$\uparrow$ &   25.292 & 27.036 & 26.874  & - & 24.132 &\textbf{27.238} \\
            & \textbf{SSIM}$\uparrow$ &    0.784 & 0.840 &  \textbf{0.863} & - & 0.734 &0.854 \\
        \midrule
        \multirow{3}{*}{QY}
            & \textbf{LPIPS}$\downarrow$  & - & - & - & - & 0.417 & \textbf{0.353} \\
            & \textbf{PSNR}$\uparrow$ &  - & - & - & - & 25.119 & \textbf{26.729} \\
            & \textbf{SSIM}$\uparrow$ &   - & - & - & - & 0.802 & \textbf{0.854} \\
        \bottomrule
        
    \end{tabular}
    \vspace{-0.1in}
\end{table*}

\begin{table}[ht]
    \centering
    \setlength{\tabcolsep}{4pt} 
    \caption{Comparison with Other Baselines.}
    \label{table:baseline}
    \begin{tabular}{c c c c c c c}
        \toprule
        \textbf{Dataset} & \textbf{Metric} &  \textbf{Splatfacto} &  \textbf{PGSR} & \textbf{GOF} & \textbf{2DGS}&\textbf{Ours} \\
        \midrule
        \multirow{3}{*}{HKU}
            & \textbf{LPIPS}$\downarrow$ & 0.197 & 0.206 & 0.281 & 0.317 &\textbf{0.173} \\
            & \textbf{PSNR}$\uparrow$ & 25.806 & 26.320 & 25.200 & 26.087 & \textbf{27.238} \\
            & \textbf{SSIM}$\uparrow$ & 0.8374 & 0.852 & 0.802 & 0.788 & \textbf{0.854} \\
        \midrule
        \multirow{3}{*}{QY}
            & \textbf{LPIPS}$\downarrow$ & 0.373 & \textbf{0.343} & 0.387 & 0.448 &0.353 \\
            & \textbf{PSNR}$\uparrow$ & 25.311 & 26.564 & 26.250 & 24.635 & \textbf{26.729} \\
            & \textbf{SSIM}$\uparrow$ & 0.844 & \textbf{0.866} & 0.854 & 0.808 &0.854 \\
        \bottomrule
    \end{tabular}
    \vspace{-0.2cm}
\end{table}

        

\vspace{2em}
\vspace{-1.1em}



\vspace{-0.1in}
\section{Conclusion} 
This work addresses the challenges of robust pose estimation and geometrically consistent 3D reconstruction in repetitive-textured outdoor scenes. We present a significant advancement in the field of 3D reconstruction by extending COLMAP's triangulation robustness with a novel optimization formulation that refines poses, introduces normal and shape constraints, and promotes coherent surface properties to ensure accurate geometric representation. Extensive experiments on public and self-collected datasets demonstrate that our method outperforms existing approaches in both accuracy and stability, particularly in large-scale environments with texture repetition. These advancements highlight the potential of our framework for practical 3D reconstruction tasks requiring rigorous geometric consistency. 
In conclusion, our method consistently outperforms other approaches across different datasets, providing a more reliable and effective solution for scenario-level real-world digital asset generation in texture-deficient outdoor scenes.









\bibliographystyle{plain}
\bibliography{root}

\end{document}